\documentclass[journal]{IEEEtran}
\usepackage{amsmath}
\usepackage{amssymb}
\usepackage{bm}
\usepackage{graphicx}
\usepackage{float}
\usepackage{caption}
\usepackage{subfigure}
\usepackage{makecell}
\usepackage{multirow}
\usepackage[numbers,sort&compress]{natbib}
\graphicspath{{/}}

\ifCLASSINFOpdf
\else
\fi
\begin{document}
%
\title{MU-GAN: Facial Attribute Editing based on Multi-attention Mechanism}
%
%
%

\author{Ke~Zhang,~\IEEEmembership{Member,~IEEE,}
	Yukun~Su,
	Xiwang~Guo,~\IEEEmembership{Member,~IEEE,}
	Liang~Qi,~\IEEEmembership{Member,~IEEE, and}
	Zhenbing~Zhao,~\IEEEmembership{Member,~IEEE,}
	\thanks{This work is supported in part by the National Natural Science Foundation of China (NSFC) under grant number 61871182, 61302163, 61401154, by Beijing Natural Science Foundation under grant number 4192055, by the Natural Science Foundation of Hebei Province of China under grant number F2015502062, F2016502101, F2017502016, by the Fundamental Research Funds for the Central Universities under grant number 2018MS094, 2018MS095, and by the Open Project Program of the National Laboratory of Pattern Recognition (NLPR) under grant number 201900051.}
	\thanks{K. Zhang,~Y. Su, and Z. Zhao are with Department of Electronic and Communication Engineering, North China Electric Power University, Hebei, 071003, P. R. China.}
	\thanks{X. W. Guo is with the Computer and Communication Engineering College, Liaoning Shihua University, Fushun, 113001, P R. China, and also with the Department of Electrical and Computer Engineering, New Jersey Institute of Technology, Newark, NJ 07102 USA. X. W. Guo is the corresponding author (e-mail: x.w.guo@163.com).}
	\thanks{L. Qi is with the College of Computer Science and Engineering at Shandong University of Science and Technology, Qingdao, 266590, P. R. China (qiliangsdkd@163.com).}
}

%
%

\markboth{IEEE/CAA JOURNAL OF AUTOMATICA SINICA,~Vol.~X, No.~X, X~X}%
{Shell \MakeLowercase{\textit{et al.}}: Bare Demo of IEEEtran.cls
for Journals}
%



\maketitle

\begin{abstract}
Facial attribute editing has mainly two objectives: 1) translating image from a source domain to a target one, and 2) only changing the facial regions related to a target attribute and preserving the attribute-excluding details. In this work, we propose a Multi-attention U-Net-based Generative Adversarial Network (MU-GAN). First, we replace a classic convolutional encoder-decoder with a symmetric U-Net-like structure in a generator, and then apply an additive attention mechanism to build attention-based U-Net connections for adaptively transferring encoder representations to complement a decoder with attribute-excluding detail and enhance attribute editing ability. Second, a self-attention mechanism is incorporated into convolutional layers for modeling long-range and multi-level dependencies across image regions. experimental results indicate that our method is capable of balancing attribute editing ability and details preservation ability, and can decouple the correlation among attributes. It outperforms the state-of-the-art methods in terms of attribute manipulation accuracy and image quality. Our code is available at \emph{https://github.com/SuSir1996/MU-GAN}.
\end{abstract}


%
\IEEEpeerreviewmaketitle

\section{Introduction}
%
%
%
%
\IEEEPARstart{F}{acial} attribute editing aims to replace some attributes of a source facial image with target attributes, such as changing a subject's hair color, gender or expression. Facial attribute editing plays an important role in human-robotics interaction and bionic agents, which has extensive applications in such fields as face reconstruction \cite{kingma2013auto}, privacy-preserving \cite{wang2018privacy} and intelligent photography \cite{he2019attgan}.

The difficulty in facial attribute editing lies in accurately manipulating a given image from a source attribute domain to a target one while keeping attribute-independent details well preserved. Facial image needs to satisfy strict geometric constraints and correlations among facial attributes. Besides, it is difficult to achieve both attribute manipulation and detail retention ability. These make facial attribute editing a difficult task. Recently, significant breakthroughs have been made with the development of Generative Adversarial Networks (GAN) \cite{liu2017unsupervised,lample2017fader,li2019global,yang2018learning,dong2017unsupervised,pumarola2018ganimation}. Some previous studies \cite{zhu2017unpaired,choi2018stargan,he2019attgan} are based on an encoder-decoder architecture, which is adopted for extracting source image representation and reconstructing it under the guidance of target attribute vectors.

\begin{figure}[t]
	\centering
	\includegraphics[width=0.5\textwidth]{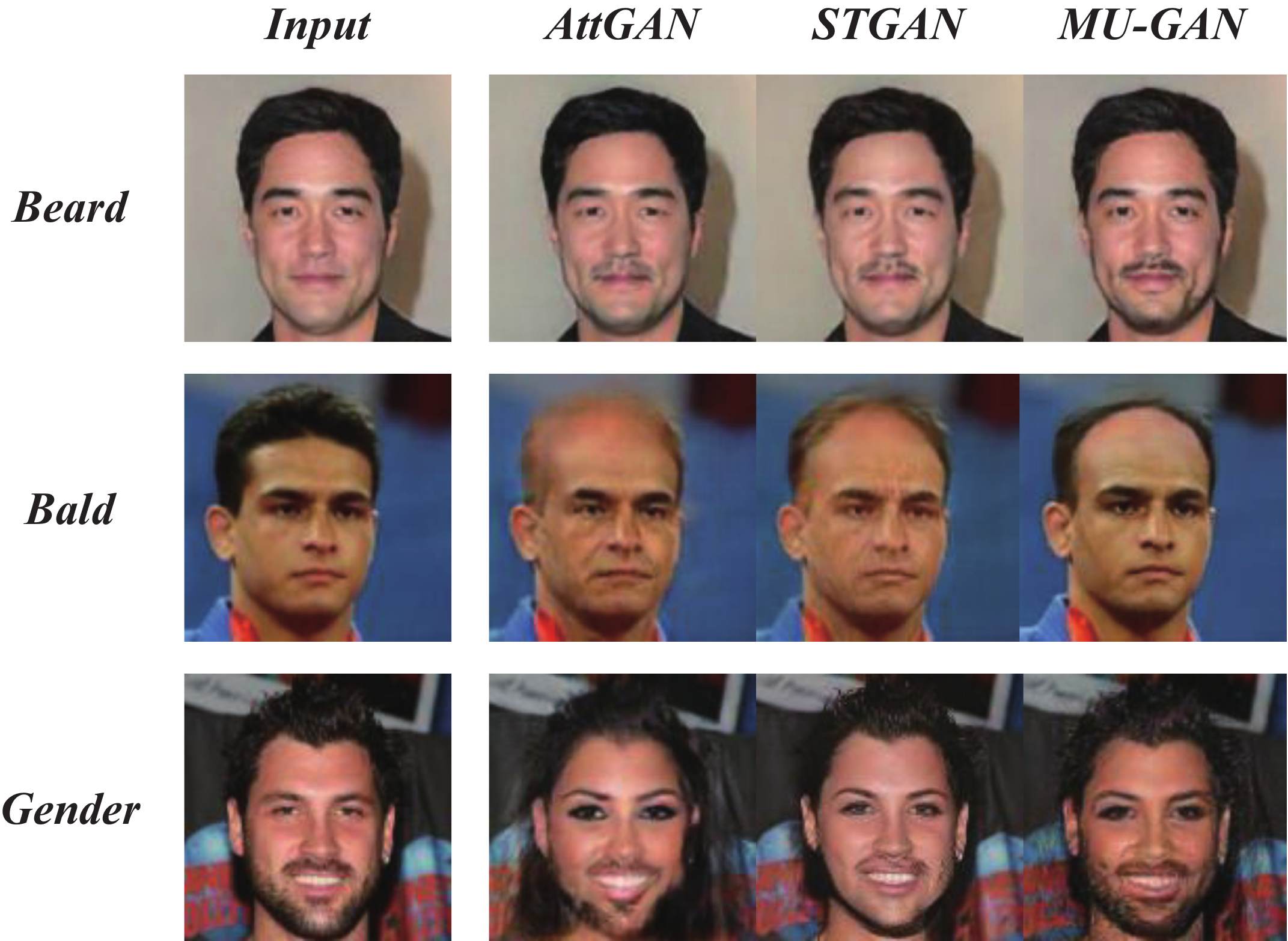}
	\caption{Image examples generated by AttGAN \cite{he2019attgan}, STGAN \cite{liu2019stgan}, and MU-GAN.}
	\label{1}
\end{figure}

Although it is widely used in image-to-image translation, an encoder-decoder architecture has some limitations especially in high quality attribute editing. Facial attributes have different levels of abstraction and can be divided into local attributes such as beard and facial aging texture, global attributes such as bald and hair color, or more abstract attributes such as gender. Convolutional downsampling or spatial pooling can be used to to obtain different levels of abstract attributes. A generator in \cite{he2019attgan} uses an encoder-decoder with residual layers \cite{he2016deep,zhang2017residual}. However, the introduction of residual bottleneck layers means that latent presentations are highly compressed and image details are thus lost during frequent down-up sampling. The innermost latent representation with minimal spatial size cannot contain all the useful details, which leads to blurry attribute-editing results and serious content-missing problems. The preservation of the details is the guarantee of image reality and quality. As a remedy, researchers \cite{ronneberger2015u} attempt to add skip-connections between an encoder and a decoder to supplement decoder representations. Encoder representations are employed as a supplement of decoder branches with detailed information. The use of direct skip-connections can transfer abundant complementary details to make images more realistic, but also transfer a lot of details, which are related to the original attributes, resulting in information redundancy and thereby weakening attribute manipulation ability. As shown in Fig. \ref{1}, the model with direct skip-connections performs bad in local attribute editing, $e.g.$, beard, with limited attribute manipulation ability. In previous studies, detail retention and attribute manipulation are difficult to reconcile.

The introduction of convolutional neural networks (CNNs) \cite{chen2018parallel}have promoted the development of GANs \cite{zhu2017unpaired,choi2018stargan,he2019attgan}. Researchers \cite{zhang2018self} believe that CNN-based GANs are good at editing local attributes and synthesizing images with few geometric constraints. Taking the landscape images as an example, a slight deformation of mountains and rivers does not affect the reality of an image. However, they have difficulty in editing images with geometric or structural patterns, such as facial attribute editing. As shown in Fig. \ref{1}, when dealing with a \emph{bald} attribute, CNN-based GANs \cite{he2019attgan,liu2019stgan} often simply paint the hair color of the original image with skin color to create the illusion of bald, ignoring the outline of a face and generating visually weird samples. One possible explanation is that CNN-based GAN relies on convolutional kernel to model global dependencies across long-range regions. Due to the limited receptive field of a convolution kernel, it is difficult to capture the dependencies among long-distance pixels in a picture.

It is also known that there are complex coupling relationships among facial attributes, $e.g.$, gender and beard. In some facial attribute editing tasks, it is unavoidable to generate samples that do not exist in the real world, such as a woman with a beard in the third row of Fig. \ref{1}. Results generated by Attribute Generative Adversarial Network (AttGAN) \cite{he2019attgan} change hair length and produces serious artifacts. Although a sample generated by Selective Transfer Generative Adversarial Network (STGAN) \cite{liu2019stgan} is more like a woman. However, it still suffers from poor attribute decoupling, which makes an undesired change in beard. Thus, a desired model needs to have the ability to decouple attributes in order to meet the requirements of target labels.

To solve these problems, we construct a new generator with a novel encoder-decoder architecture and propose a Multi-attention U-Net-based GAN (MU-GAN) model.
First, for detail preservation, a symmetric U-Net architecture \cite{ronneberger2015u} is employed to replace the original asymmetric one to ensure that the abstract semantics of latent representations at both sides of an encoder-decoder are in the same level, and avoid the information loss caused by sharp decrease in channel count numbers of the last decoder layer. Second, an additive attention mechanism is introduced to U-Net skip-connections, so that attribute-excluding representations are selectively transferred under the guidance of an attention mask, which is complementary to decoder representations and helps us balances detail preservation and attribute manipulation abilities. Third, self-attention layers are introduced to an encoder-decoder as a supplement to the convolutional layers. A self-attention mechanism helps us model long-range dependencies among image regions. It can effectively capture multi-level presentations and help GAN enforce complicated geometric constraints on generated images. In addition, the use of a multi-attention mechanism makes the model more powerful in attribute decoupling.

our method is capable of generating facial images with better perception reality, attribute manipulation accuracy, and geometric rationality, compared with the state-of-the-art approaches. Moreover, the new generator architecture can balance attribute manipulation and detail preservation abilities. As shown in Fig. \ref{1}, our model performs well in attribute editing tasks at different semantic levels with strong attribute decoupling capability. In summary, this work makes the following contributions:

• It constructs a symmetric U-Net-like architecture generator based on an additive attention mechanism, which effectively enhances our method's detail preservation and attribute manipulation abilities.

• It takes a self-attention mechanism into the existing encoder-decoder architecture thus effectively enforcing geometric constraints on generated results.

• It introduces a multi-attention mechanism to help attribute decoupling, $i.e.$, it only changes the attributes that need to be changed. Qualitative and quantitative results show that MU-GAN outperforms the state-of-the-art methods in facial attribute editing.

The rest of the paper is organized as follows. Section II briefly reviews related work for generative model, image-to-image translation and facial attribute editing. The proposed method is illustrated in Section III. Experimental results and analysis are presented in Section IV. Ablation study is described in Section V, leading to conclusions in Section VI.

\section{Related Work}
\subsection{Generative Model}
A generation model is devoted to learning real sample distribution and have attracted upsurging attention in attribute editing. There are two main approaches for facial generation models: variational auto-encoder (VAE) \cite{kingma2013auto} and GAN. The former's goal is to maximize variational lower bounds, while GAN aims to reach Nash equilibrium through a binary mini-max game. experimental results show that VAE's training process is more stable, but the results are fuzzy. GAN has better generation quality and creativity than VAE, but lacks appropriate constraints. To address the above issues, Wasserstein Generative Adversarial Networks (WGANs) \cite{arjovsky2017wasserstein,gulrajani2017improved} improve stability of the optimization process by replacing \emph{Jensen-Shannon/Kullback–Leibler} divergence \cite{wang2019drifted} with \emph{Earth-Mover} distance to measure the distance between real and generated sample distribution, thus solving the problem of vanishing gradient. A conditional image generation task has also been actively studied. Several methods \cite{mirza2014conditional,odena2016semi,odena2017conditional,reed2016generative,zhang2017stackgan,shu2017neural,taigman2016unsupervised,kim2017learning,ledig2017photo,xu2019adaptive,gao2018dendritic,guo2019lexicographic,guo2017dual} use category information such as attribute labels to generate samples. GAN \cite{goodfellow2014generative,wang2017generative} has exhibited a remarkable capability in various fields, and has been used in several applications such as image generation \cite{goodfellow2014generative,arjovsky2017wasserstein,gulrajani2017improved,qi2019loss}, style translation \cite{liu2016coupled,lample2017fader,almahairi2018augmented,choi2018stargan,li2019global}, super-resolution, image reconstruction \cite{kingma2013auto,xiang2019single}, and facial attribute editing \cite{li2019global,yang2018learning,dong2017unsupervised,pumarola2018ganimation,zhou2017genegan,xiao2017dna,perarnau2016invertible,he2019attgan,liu2019stgan}.
\subsection{Image-to-Image Translation}
Image-to-image translation, $i.e.$, manipulating a given image attribute from a source domain to a target one with other image contents untouched. Existing works \cite{isola2017image,liu2017unsupervised,liu2015deep,zhu2017unpaired,guo2015disassembly,cai2019unsupervised} have made remarkable progress in image translation. For example, pix2pix \cite{isola2017image} adapts Conditional Generative Adversarial Network (CGAN) \cite{mirza2014conditional} for multi-domain image-to-image translation tasks with paired images. However, paired image datasets are unavailable in most scenarios. To address this issue, researchers \cite{kim2017learning,liu2017unsupervised,zhu2017unpaired} propose unpaired image translation methods. Unsupervised Image-to-image Translation Networks (UNIT) \cite{liu2017unsupervised} combines VAE \cite{kingma2013auto} and Coupled Generative Adversarial Network (coGAN) \cite{liu2016coupled} to build a GAN architecture, where two generators share the same weights to learn the joint distribution of images in cross domains. Cycle-Consistent Generative Adversarial Network (CycleGAN) \cite{zhu2017unpaired} preserves the key representation between the input and generated images by minimizing cycle consistency loss. The idea of dual learning allows Disco Generative Adversarial Network (DiscoGAN) \cite{kim2017learning} and CycleGAN \cite{zhu2017unpaired} to learn reversible mapping among different domains in unpaired image-to-image translation. However, the aforementioned methods cannot perform image manipulation on multiple domains. Their inefficiency results from the fact that in order to learn all mappings among $k$ domains, $k\times(k-1)$ generators have to be trained. Recent studies \cite{lample2017fader,almahairi2018augmented,choi2018stargan} focus on multi-domain conversion and propose some multi-domain image translation models such as Augmented CycleGAN \cite{almahairi2018augmented}, Star Generative Adversarial Network (StarGAN) \cite{choi2018stargan}, and AttGAN \cite{he2019attgan}.
\subsection{Facial attribute editing}
The objective of facial attribute editing is to generate a face with a target attribute while preserving the attribute-excluding facial detail. Facial attribute editing has been a hot topic in computer vision. Existing methods \cite{li2019global,yang2018learning} are designed for modeling an aging process. Face aging is mainly reflected by wrinkles. Since the subtle texture information is more salient and robust in a frequency-domain, a Wavelet-domain Global and Local Consistent Age Generative Adversarial Network (WaveletGLCA-GAN) \cite{li2019global} uses wavelet transform to synthesize aging images. Several studies \cite{dong2017unsupervised,pumarola2018ganimation,hu2018emotion} are conducted to solve a facial expression synthesis problem. Other studies propose facial attribute editing methods. DNA Generative Adversarial Network (DNA-GAN) \cite{xiao2017dna} can be regarded as an extension of Gene Generative Adversarial Network (GeneGAN) \cite{zhou2017genegan}, which swaps attribute-relevant latent representations between given image pairs to synthesize "hybrid" images, and can transform multiple attributes simultaneously. Fader Network (FaderNet) \cite{lample2017fader} imposes adversarial constraints to enforce the independence of latent representations. Its decoder then takes latent representation extracted from an encoder and target attribute vector as the input to generate desired results. Invertible conditional Generative Adversarial Network (IcGAN) \cite{perarnau2016invertible} and FaderNet \cite{lample2017fader} impose the constraints of mutual independence of attributes on the  latent space such that latent representations from different classes can be independent of attribute decoupling. On the contrary, experimental results \cite{he2019attgan} prove that it is too strict to impose independent constraints on latent space. Then, AttGAN \cite{he2019attgan} applies attribute classification constraints to generated images to ensure attributes being translated correctly. The generator of AttGAN consists of five convolution and deconvolutional layers. Then, it applies one skip-connection between encoder and decoder to improve image quality. Note that AttGAN's encoder-decoder is not a symmetrical structure, and sharp decrease in the number of channels in the last deconvolutional layer of its decoder results in detail loss. Limited by the receptive field of a convolution kernel, CNN layers cannot model long-range, multi-level dependencies across image regions, which makes it difficult to synthesize image classes with complex geometric or structural patterns. Previous work \cite{ronneberger2015u} adopts skip-connections to enhance detail retention at the cost of reducing attribute manipulation ability. Adding direct skip-connections cannot fundamentally balance the attribute manipulation and detail retention abilities. AttGAN and its variants face three problems: 1) Loss of image details; 2) Insufficient attribute manipulation ability; and 3) Poor enforcement of geometric constraints. STGAN \cite{liu2019stgan}, a variant of AttGAN, introduces Gated Recurrent Unit (GRU) \cite{cho2014learning} to build selective transfer units to selectively transmit encoder representation. However, memory-based approaches, $e.g.$, GRU \cite{cho2014learning} and Long Short-Term Memory (LSTM) \cite{hochreiter1997long,wang2017generative,principi2019unsupervised,zhang2019fine} mainly focus on sequential processing rather than visual tasks, which is limited by memory capacity and low computational efficiency.

\section{Proposed Method}
\begin{figure}[htpb]
	\centering
	\includegraphics[width=0.5\textwidth]{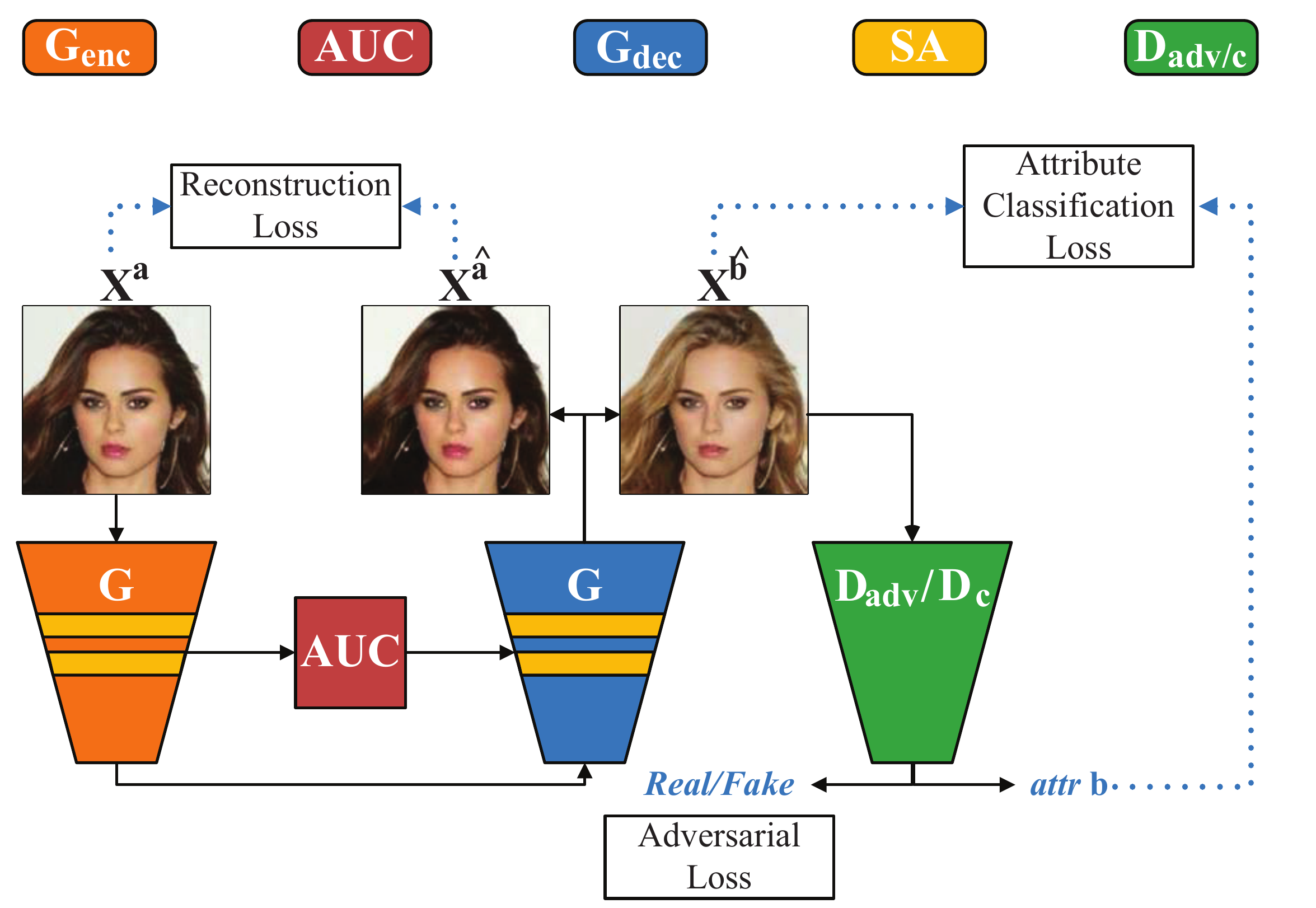}
	\caption{The architecture of MU-GAN. SA denotes a self-attention mechanism. MU-GAN consists of a generator $G$ and a discriminator $D$. $D$ consists of two sub-networks, $i.e.$, real/fake adversarial discriminator $D_{adv}$ and Attribute Classifier $D_c$, which share the weights of the same convolutional layers. AUCs bridge an encoder $G_{enc}$ and a decoder $G_{dec}$ to selectively transform encoder presentation, making it complementary to decoder presentation.}
	\label{2}
\end{figure}

Fig. \ref{2} shows an overview of our method. In order to solve the problem of AttGAN and STGAN, we present MU-GAN for facial attribute editing. First, instead of using an ordinary encoder-decoder \cite{he2019attgan}, we use a symmetric U-Net structure to build our generator and construct MU-GAN by replacing direct skip-connections with attention U-Net connections (AUCs). Second, we adopt self-attention layers as complement to convolution layers. Finally, a discriminator and objective function of MU-GAN are provided.

\begin{figure*}[htpb]
	\centering
	\includegraphics[width=\textwidth]{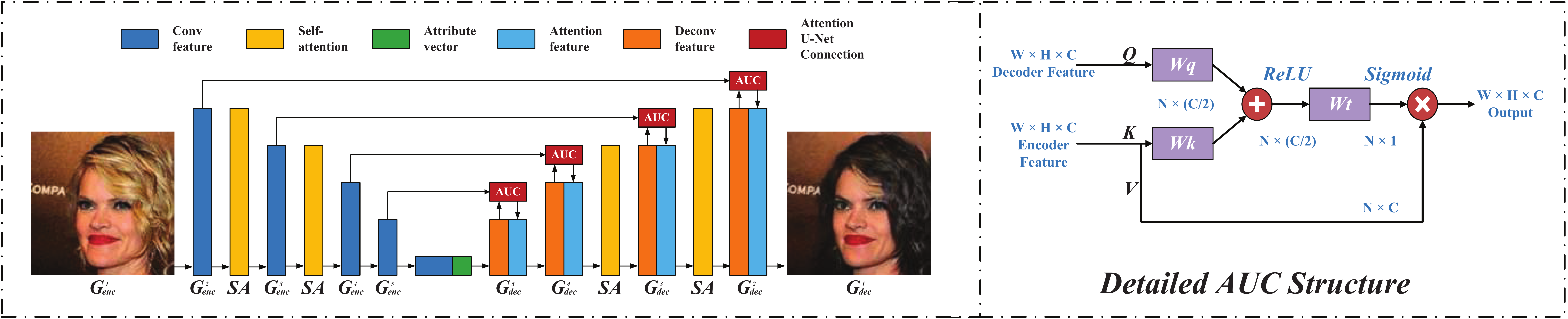}
	\caption{The architecture of the proposed generator. AUCs bridge the two ends of an encoder-decoder, and calculates attention coefficient $\alpha$ between encoder-decoder representations of the same size. Under the guidance of $\alpha$, it selectively transfers encoder representation as supplementary information of a decoder. Green block represents target vectors used to guide attribute editing. Besides, we follow SAGAN and put self-attention layers behind convolutional layers with feature map sizes of 64 and 32, respectively.}
	\label{3}
\end{figure*}

\subsection{Generator}
\subsubsection{Attention U-Net Connection}
Fig. \ref{3} shows the architecture of the proposed generator and AUCs. For detail retention and blurry image problems, we replace the original asymmetric CNN-based encoder-decoder with a symmetrical Attention U-Net architecture. Besides, instead of directly connecting an encoder to a decoder via skip-connections, we present AUCs to selectively transfer attribute-irrelevant representations from an encoder, and then, AUCs concatenate encoder representation with decoder ones to improve image
quality and detail preservation. With an attention mechanism, AUCs are capable of filtering out representations related to original attributes while preserving attribute-irrelevant details. It can promote the image fidelity without weakening attribute manipulation ability. By using an attention mechanism, AUCs solve the problem of information redundancy caused by direct skip-connections.

We modify the additive attention mechanism \cite{vaswani2017attention} to build AUCs. Like other classic encoder-decoder attention mechanisms, Query ($Q$) comes from decoder representation $\bm{d} \in \mathbb{R}^{C \times H \times W}$. Meanwhile, Key ($K$) along with Value ($V$) consist of encoder representations $\bm{e} \in \mathbb{R}^{C \times H \times W}$.

Without loss of generality, we take encoder/decoder layers $\bm{l}$ as an example. Image representations $\bm{e}^l$/$\bm{d}^l$ from the previous encoder/decoder layers are first transformed into two feature spaces $\bm{q}$ and $\bm{k}$ $\in \mathbb{R}^{C \times N}$, through independent linear transformations, $i.e.$, $W_q$ and $W_k$. Note that $N=W \times H$. Representations are reshaped to be a vector $(W \times H)\times(C/2)$. Let $i$ denotes the $i$-th position in a vector. The linear transformations are implemented by a channel-wise 1$\times$1 convolutions, and the number of representation channels is reduced to $C/2$, $i.e$, half of the input size.

\begin{gather}
\bm{q(d_i^l) = W_q^{\top}d_i^l} \text{,\quad}\bm{k(e_i^l) = W_k^{\top}e_i^l}
\end{gather}

Additive similarity $\bm{a}_i^l$ is obtained by adding $\bm{q(d_i^l)}$ and $\bm{k(e_i^l)}$. Then, through an activation function $ReLU$, another transformation block $\bm{W_t}$ and $Sigmoid$ functions, attention map $\bm{\alpha}$ is calculated, which denotes as:
\begin{gather}
\bm{a}_i^l=ReLU(\bm{q}(\bm{d}^l_i)+\bm{k}(\bm{e}^l_i))\\
\bm{\alpha_i}=\frac{1}{1+\exp{(-\bm{W}_t^{\top}\bm{b}_i^l)}}
\end{gather}
Where, attention coefficient, $\bm{\alpha}_i \in [0,1]$, identifies salient image regions and prune representation in order to preserve only the activation of attribute-excluding information.

The output of AUCs is an element-wise multiplication of encoder representation $\bm{e}^l_i$ and attention coefficients $\bm{\alpha}_{i}$, $i.e.$,
\begin{gather}
\bm{\hat{e}}^l=\sum_{i=1}^{N}\bm{\alpha}_{i}^l \bm{e}^l_i
\end{gather}

Finally, as shown in Fig. \ref{3}, encoder representation $\bm{\hat{e}}^l$ transferred by AUCs is concatenated with decoder representation $\bm{d_i^l}$, which is of the same scale. This combined representation is an input of the subsequent upsampling layer.

AUCs progressively suppress representation responses in source-attribute-related regions, and retains image details that are independent of the attributes. Representations transferred by AUCs are used as supplementary to decoder representations to compensate for the irreversible information loss caused by convolution downsampling, and enrich the details of a concerned image.

More importantly, as shown in Fig. \ref{3}, AUCs help $G$ aggregate information from multiple image scales, which increases the image fidelity and achieves better performance, without weakening attribute manipulation ability.

Note that our method adopts a symmetrical encoder-decoder to settle the issue of highly-compressed representation and loss of details caused by the sharp decrease in the number of channels. In addition, the abstract-level of representations at both ends of a symmetric encoder-decoder are similar, which are highly correlated with each other and contain significant reference values for attribute editing.

\subsubsection{self-attention}
Most GAN-based models for facial attribute editing are built with convolutional layers. Limited by the receptive field of a convolution kernel, the convolutional layer can only process information from adjacent pixels. Therefore, many CNN-based GAN models share similar problems, $i.e.$, their results poorly meet global geometric constraints, and the networks are not competent for an image manipulation task with complex composition and strict geometric constraints.

\begin{figure}[h]
	\centering
	\includegraphics[width=0.5\textwidth]{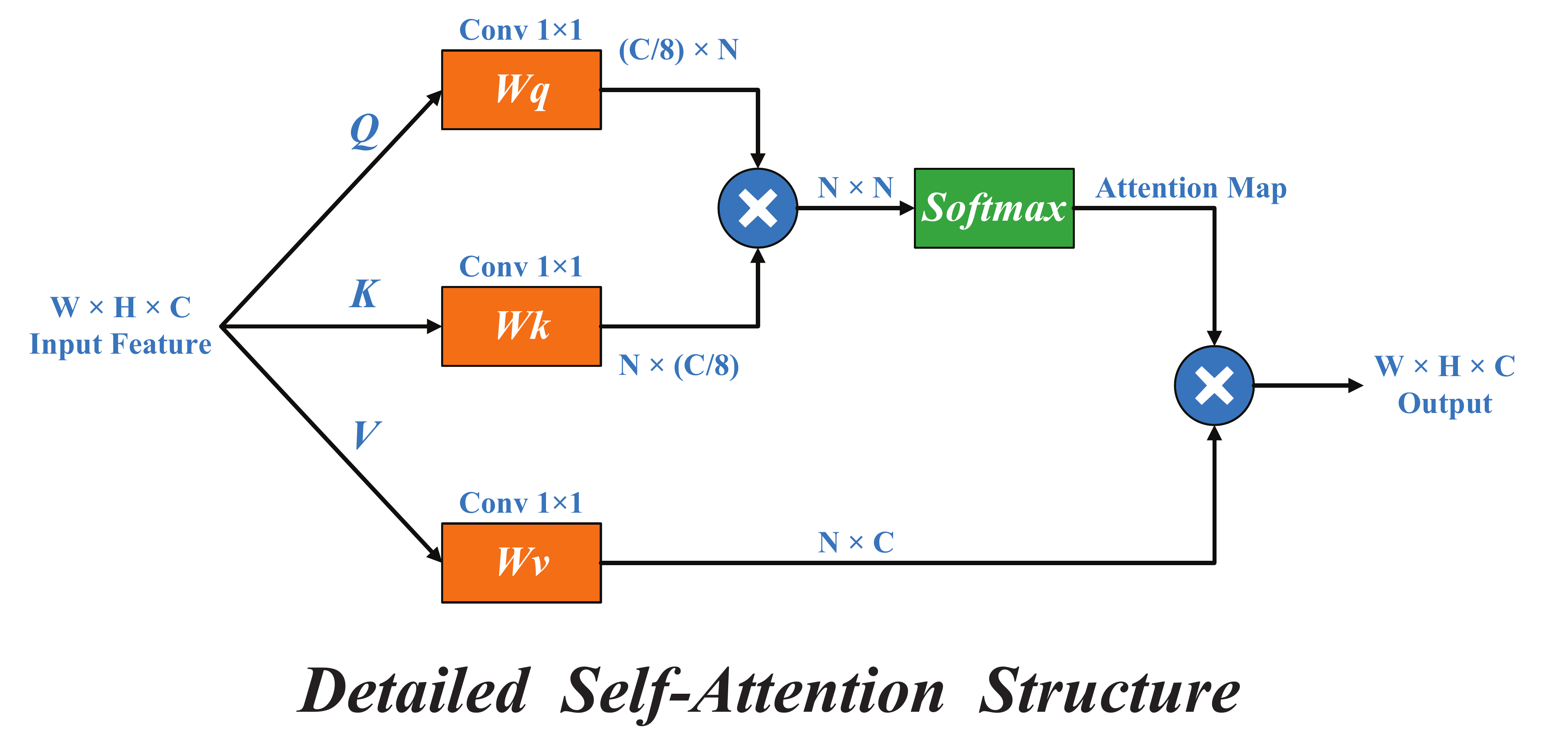}
	\caption{Structure of a self-attention mechanism, where $\otimes$ represents matrix multiplication. After the features pass through $W_q$, $W_k$ and $W_v$, their feature size is reshaped. Note that $N=W \times H$.}
	\label{4}
\end{figure}

For example, the task of facial attribute editing requires rigorous arrangement of facial features, and a tiny unreasonable-deformation can cause salient visual irrationality. As shown in the 2nd row of Fig. \ref{1}, the edited results fail to meet appropriate structural and geometric constraints. Thus, we utilize a self-attention mechanism \cite{zhang2018self,wang2018non} as a supplement to convolutional layers in $G$, to efficiently model dependency across long-range separated spatial regions. The details of a self-attention mechanism are shown in Fig. \ref{4}.

In a self-attention layer, representation $\bm{x\in \mathbb{R}^{C{\times}W{\times}H}}$ extracted from the previous CNN layer is employed as the input of $Q$ and $K$. $\bm{x}$ is transformed into two representations spaces, $\bm{q \in \mathbb{R}^{(C/8){\times}N}}$ and $\bm{k \in \mathbb{R}^{(C/8){\times}N}}$, to calculate attention coefficient $\bm{\beta}$, where $\bm{q(x) = W_q^{\top}x}$ and $\bm{k(x) = W_k^{\top}x}$.
\begin{gather}
\bm{b}_{ij}=\bm{k}(x_i)^{\top}\bm{q}(x_j) \\
{\bm{\beta}_{j,i}}=\frac{\exp{\bm{b}_{ij}}}{\sum_{i=1}^{N}\exp{\bm{b}_{ij}}}~~~~\forall i,j \in [1,N]
\end{gather}

The softmax operation is performed on each row. The output is an element-wise multiplication of encoder representation $\bm{v(x)}$ and attention coefficient $\bm{\beta}$. Finally, the output is given as:
\begin{gather}
\bm{o}_j=\sum_{i=1}^{N}\bm{v(x_i)}\bm{\beta}_{j,i}
\end{gather}
where $\bm{v(x)=W_v^{\top}x}$.

\subsection{Discriminator}
Discriminator $D$ consists of two sub-networks: $real$/$fake$ adversarial discriminator $D_{adv}$, and facial attribute classifier $D_c$. Its backbone consists of five stacking convolutional layers used to capture the feature of an input image. All the convolutional layers use instance normalization and $leaky \text{\,} ReLU$ function. After the last convolutional layer, the CNN backbone is divided into two branches, which are connected with two independent fully-connected layers $D_{adv}$ and $D_{c}$, respectively. While $G$ is trying to generate an image $G(\bm{x}^a,\bm{b})$ conditioned on both target label $\bm{b}$ and input facial image $\bm{x}^a$, $D_{adv}$ aims to distinguish $\bm{x}^{\hat{b}}$ from real images and $D_c$ attempts to verify if the output image contains desired attributes.

\subsection{Loss Functions}
In MU-GAN, generator $G$ consists of two sub-networks $G_{enc}$ and $G_{dec}$. The former encodes an input image into a latent representation, while the latter recovers it under the guidance of target attribute label $\bm{b}$, to transfer image $\bm{x}^a$ from a source domain to a target one.
Given a face image $\bm{x}^a$ with $\bm{n}$ binary attribute label $\bm{a}$, $G_{enc}$ extracts its latent representation through five convolutional layers, defined as:
\begin{gather}
\bm{F}_{e} = G_{enc}(\bm{x}^a) \\
\bm{F}_{e} =\text{\{}\bm{f}^1_{e} \text{…} \bm{f}^5_{e}\text{\}}
\end{gather}
where $\bm{F}_{e}$ refers to output encoder representations.

Taking layer $\bm{i}$ of an encoder-decoder as an example. $\bm{f}^i_{e}$/$\bm{f}^i_{d}$ refer to output representations extracted through encoder/decoder layer $\bm{i}$, and $\bm{f}^i_{in}$ denotes input representation of decoder layer $\bm{i}$. Then, we concatenate the innermost encoder representations $\bm{f}^5_{e}$ with target attribute vector $\bm{b}$ and send it into a decoder. Guided by attribute $\bm{b}$, AUCs are deployed to transfer encoder representations for each decoder layer.
\begin{gather}
\tilde{\bm{f}}^{i-1}_{e} = AUC(\bm{f}^{i-1}_{d}, \bm{f}^{i-1}_{e}) \\
\bm{f}^{i-1}_{in} = \mathbb{C}(\tilde{\bm{f}}^{i-1}_{e}, \bm{f}^{i-1}_{d}) \\
\bm{f}^i_{d} = \mathbb{D}(\bm{f}^{i-1}_{in})
\end{gather}
where $\mathbb{C}$, $\mathbb{D}$ denote channel-wise concatenation function and deconvolutional layers, respectively. $\bm{f}^{i-1}_{e}$ is selectively transferred from encoder by an additive attention mechanism, and then we concatenate it with $\bm{f}^{i-1}_{dec}$ as input representation, $i.e.$, $\bm{f}^{i-1}_{in}$ for the next transpose convolution layer. Finally, through $G_{enc}$ and $G_{dec}$, $\bm{x}_a$ is transformed into a new image $\bm{x}^{\hat{b}}$ with target attributes:
\begin{gather}
\bm{x}^{\hat{b}} = G_{dec}(G_{enc}(\bm{x}^a),\bm{b})=G(\bm{x}^a,\bm{b})
\end{gather}

Next, we cover the adversarial, attribute classification and reconstruction losses.

\subsubsection{Adversarial Loss}
\begin{figure*}[tpb]
	\centering
	\includegraphics[width=\textwidth]{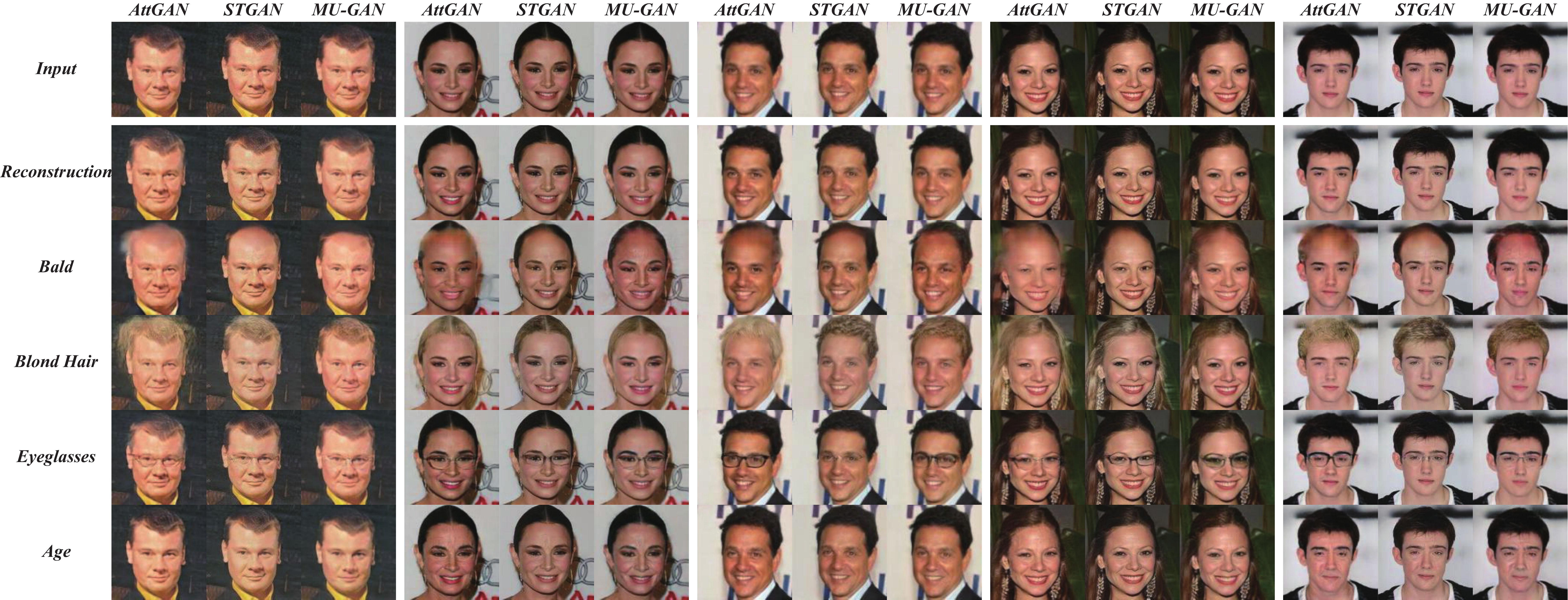}
	\caption{Comparisons with AttGAN \cite{he2019attgan} and STGAN \cite{liu2019stgan} on editing specified attributes.}
	\label{5}
\end{figure*}

In order to make the distribution of generated images close to the distribution of real images, we introduce adversarial learning in the proposed method, thereby improving the visual reality of the generated image.
WGAN uses $Earth-Mover$ distance as a metric to measure the distance between two probability distributions, which can make the training process more stable and avoid mode collapse from happening. Following WGAN, we formulate adversarial loss between $G$ and $D_{adv}$ as follows:
\begin{gather}
\min_{D_{adv}}\mathcal{L}_{adv}=-\mathbb{E}_{x^a{\sim}P_{r}}D(x^a)+\mathbb{E}_{x^{\hat{b}}{\sim}P_f}D(x^{\hat{b}}) \\
\min_{G}\mathcal{L}_{adv}'=-\mathbb{E}_{x^{\hat{b}}{\sim}P_f}D(x^{\hat{b}})
\end{gather}
where, $P_{r}/P_{f}$ denote the real/fake image sample distribution, respectively.
\subsubsection{Attribute Classification Loss}
In order to precisely transfer image $\bm{x}^a$ into image $\bm{x}^{\hat{b}}$ with target facial attribute $\bm{b}$, attribute classifier $D_c$ is used for facial attribute classification, which imposes attribute constraints on $G$ to generate an image with correct facial attributes. Attribute classification losses are defined as:
\begin{gather}
\min_{D_c}\mathcal{L}_{cls} = \mathbb{E}_{x^{a} \sim P_{r}}[\mathbb{L}(x^a,a)]
\end{gather}
\begin{equation}
\begin{aligned}
\mathbb{L}(x^a,a) = -\sum_{i=1}^n[ & a_i\log{(D_c^i(x^a))}\\ & +(1-a_i)\log{(1-D_c^i(x^a))}]
\end{aligned}
\end{equation}
and that for $G$ is:
\begin{gather}
\min_{G}\mathcal{L}_{cls}' = \mathbb{E}_{x^{\hat{b}} \sim P_{f}}[\mathbb{L}(x^{\hat{b}},b)]
\end{gather}
\begin{equation}
\begin{aligned}
\mathbb{L}(x^{\hat{b}},b)=-\sum_{i=1}^n[ & b_i\log{(D_c^i(x^{\hat{b}}))}\\ & +(1-b_i)\log{(1-D_c^i(x^{\hat{b}}))}]
\end{aligned}
\end{equation}
where, $\bm{n}$ represents the number of attribute categories. $D_{c}^i$ represents the predicted label of the $\bm{i}$-th attribute of $D_{c}$. $\mathbb{L}$ denotes the summation of binary cross entropy loss of all the attributes.

\subsubsection{Reconstruction Loss}
The use of adversarial and classification losses cannot guarantee that only the attribute relevant regions are changed, while attribute-excluding details are well preserved. Therefore, $G_{dec}$ is required to learn to reconstruct image $\bm{x}^a$ from the $G_{enc}$ latent representations $\bm{F}_{enc}$ under the condition of original attribute label $\bm{a}$. $\bm{L}_1$ norm is introduced as a criterion to measure the similarity between generated image $x^{\hat{\bm{a}}}$ and the original image $\bm{x}^a$. The reconstruction loss is defined as:
\begin{gather}
\mathcal{L}_{rec}=\mathbb{E}_{X_a \sim P_{r}}\|x^a-G(x^a,a)\|_1
\end{gather}
where subscript 1 represents $\bm{L}_1$ norm, that performs better than $\bm{L}_2$ norm in suppressing blurriness.

\subsubsection{Overall objective}
Combining all the loss functions mentioned above, our method possesses both attribute editing and detail retention abilities. The objective for the $G$ is formulated as:
\begin{gather}
\min_{D}L=\mathcal{L}_{adv} + \lambda_1\mathcal{L}_{cls}
\end{gather}
and for $G$ it is:
\begin{gather}
\min_{G}L'=\mathcal{L}_{adv}' + \lambda_2\mathcal{L}_{cls}' + \lambda_3\mathcal{L}_{rec}
\end{gather}
where $\lambda_1$-$\lambda_3$ are the hyper-parameters of the loss function.
\section{Experiments}
\begin{figure*}[htpb]
	\centering
	\includegraphics[width=\textwidth]{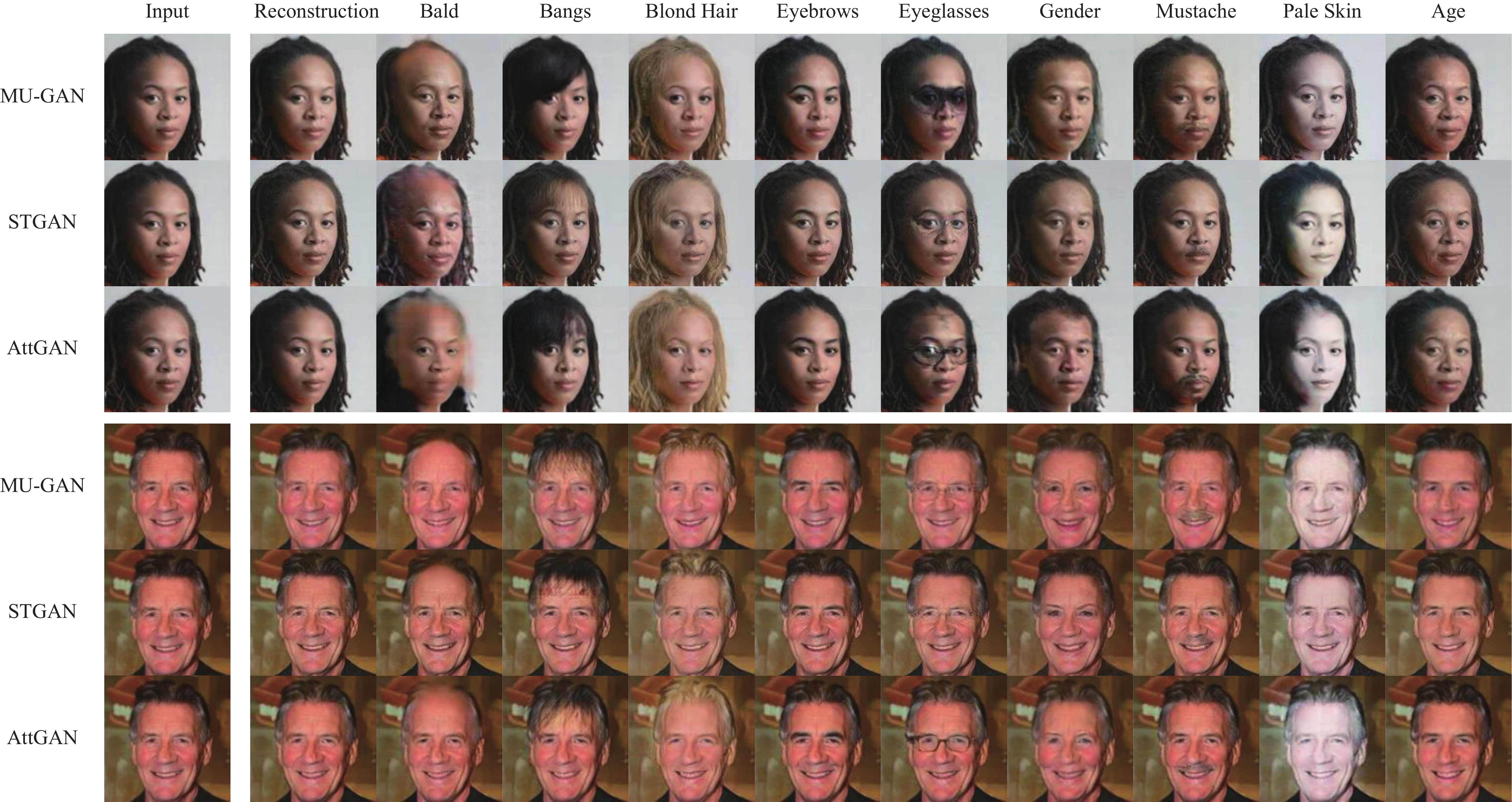}
	\caption{Facial attribute editing results of AttGAN \cite{he2019attgan}, STGAN \cite{liu2019stgan} and MU-GAN. Please zoom in for better observation.}
	\label{6}
\end{figure*}
\subsection{Implementation Details}
To evaluate the proposed method, we compare MU-GAN with AttGAN \cite{he2019attgan} and STGAN \cite{liu2019stgan} and conducted extensive experiments on the CelebA dataset. The models involved in the experiment are trained on a workstation equipped with an Intel (R) Xeon (R) CPU E5-2620 v4 @ 2.10GHz and NVIDIA GTX1080ti GPU. All the experiments are conducted in the Pytorch 0.4 environment, with Cuda 8.0.44 and cuDNN6.0.20. The baseline model is trained under original experimental setting. There are 100 epochs in the training phase and models are trained by $Adam$ optimizer ($\beta_1$ = 0.5, $\beta_2$ = 0.999), and the initial learning rate is 0.002, which drops to 1/10 of itself for every 33 epochs. We use 5 discriminator update steps per generator update during training. The weights of the objective function are set as $\lambda_1$ = 3, $\lambda_2$ = 10, and $\lambda_3$ = 100.

CelebA is a large-scale facial attributes dataset including 10,177 celebrities and 202,599 facial images, each of which have 40 binary attribute labels. In order to compare with the previous work \cite{he2019attgan,liu2019stgan}, the same data preprocessing method is adopted. Thirteen attributes with intense visual impact are selected, including \emph{Bald}, \emph{Bangs}, \emph{Black Hair}, \emph{Blond Hair}, \emph{Brown Hair}, \emph{Bushy Eyebrows}, \emph{Eyeglasses}, \emph{Male}, \emph{Mouth Slightly Open}, \emph{Mustache}, \emph{No Beard}, \emph{Pale Skin} and \emph{Young}. These attributes cover most distinctive facial attributes, containing practical information about human-computer interaction, and are also widely used in relevant work \cite{he2019attgan,liu2019stgan}. In this experiment, CelebA  source images with a size of 178$\times$218 are center-cropped and resized to 128$\times$128. According to the official division, CelebA is divided into a training set, a validation set, and a test set. The training and validation sets are used to train our method, while the test set is used in the evaluation phase.

\subsection{Qualitative results}
The qualitative results are shown in Figs. \ref{5}-\ref{6}. Some samples generated by AttGAN and STGAN suffer from low-quality problems, $i.e.$, artifacts and blurry to some extent, while the results of our method are more natural and realistic. MU-GAN aims to change only the facial attributes that need to be changed. The performance of detail preservation ability can be evaluated in two aspects. One is the preservation of details in the visual spatial regions, which is mainly reflected by whether the model can distinguish the attribute-relevant/irrelevant regions. The other one is the ability to disentangle attributes in abstract semantics. As we know, some attributes are highly correlated with other attributes, which may lead to undesired changes in other attributes.

\begin{figure*}[htbp]
	\centering
	\subfigure[Local attributes, $e.g.$ \emph{Mustache} and \emph{Beard}.]{
		\begin{minipage}[t]{0.33\textwidth}
			\centering
			\includegraphics[width=\textwidth]{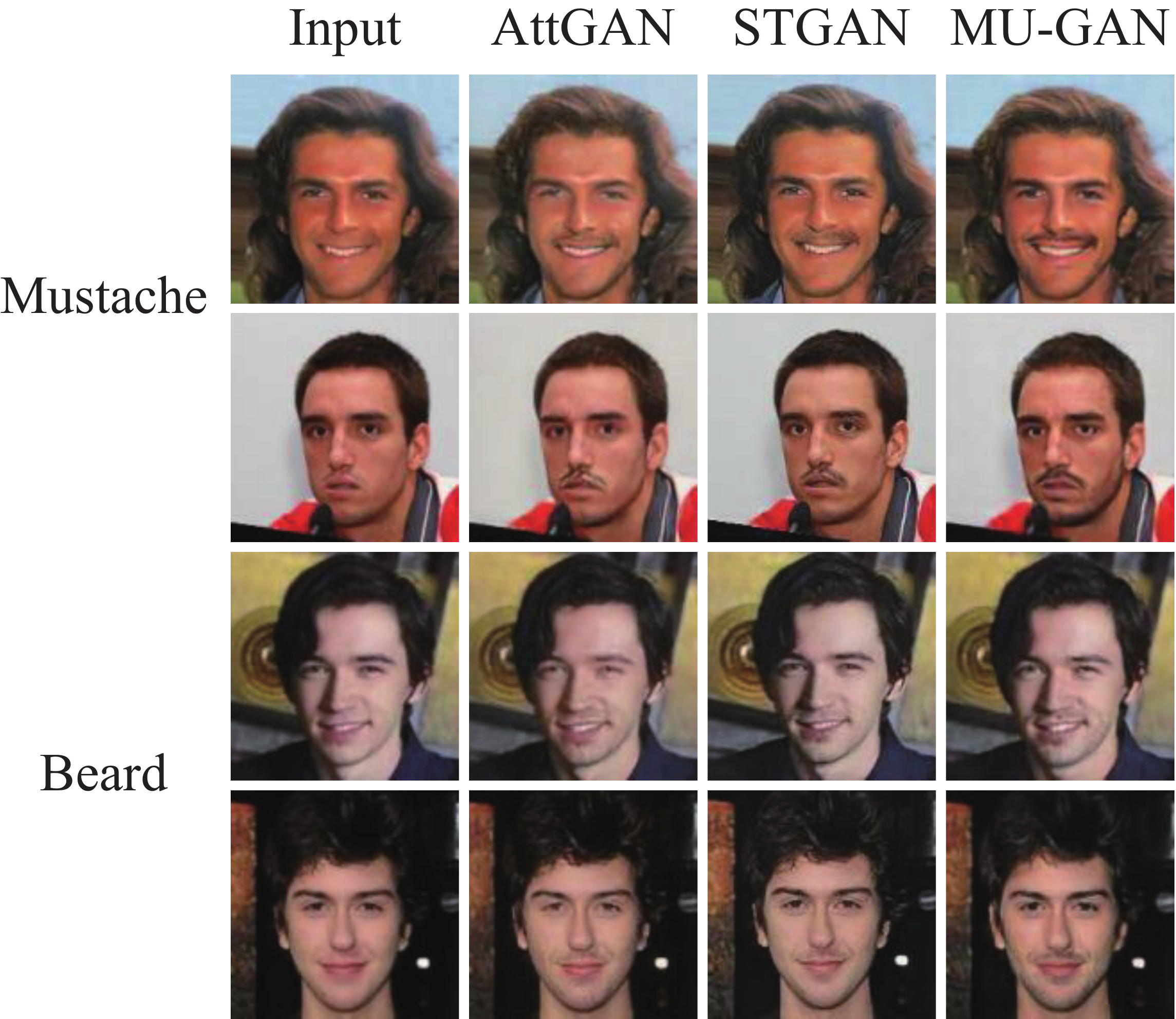}
			\label{7-a}
		\end{minipage}%
	}%
	\subfigure[Global attributes, $e.g.$ \emph{Bald} and \emph{Skin}.]{
		\begin{minipage}[t]{0.33\textwidth}
			\centering
			\includegraphics[width=\textwidth]{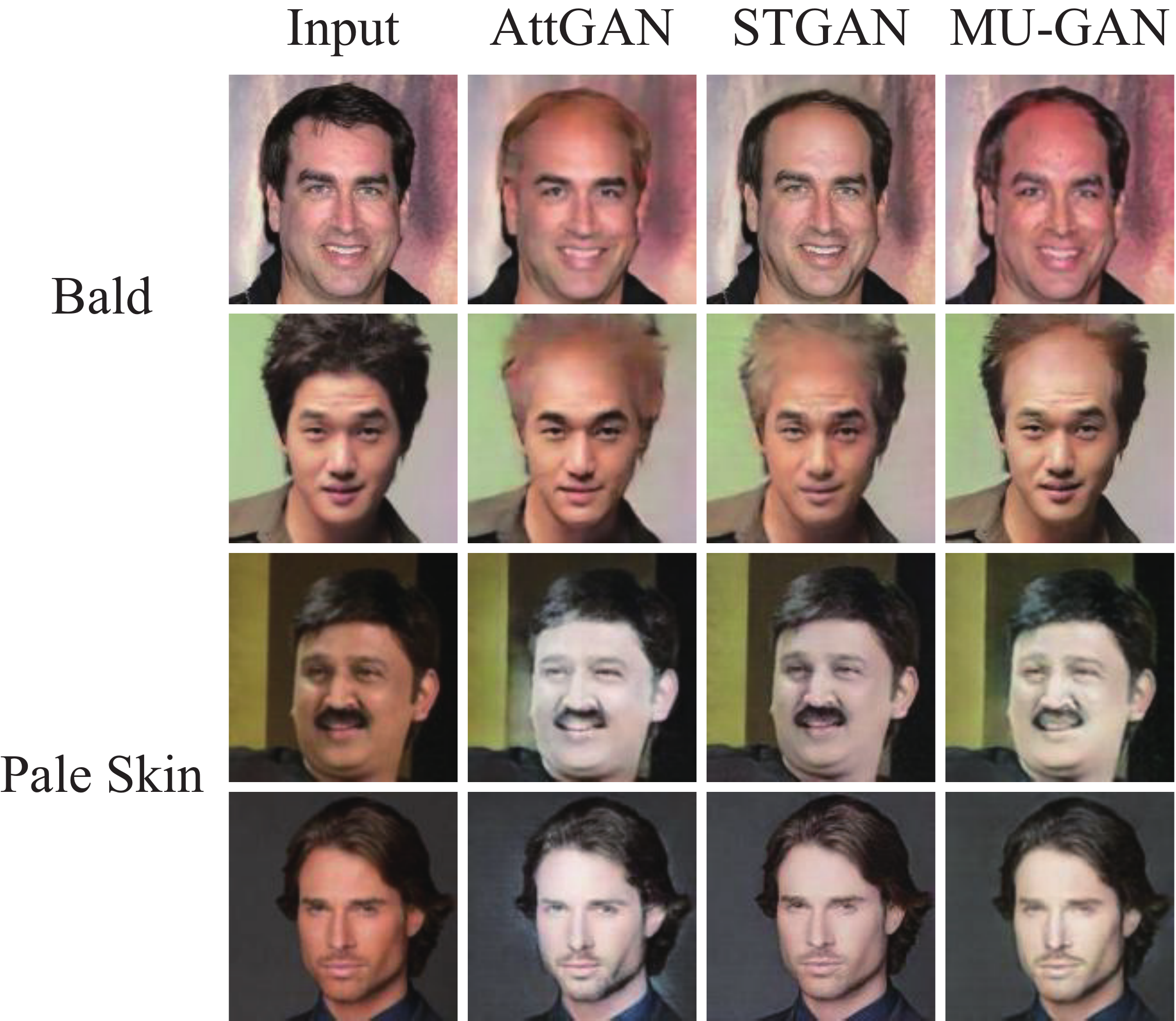}
			\label{7-b}
		\end{minipage}%
	}%
	\subfigure[Abstract attributes, $e.g.$ \emph{Gender}.]{
		\begin{minipage}[t]{0.33\textwidth}
			\centering
			\includegraphics[width=\textwidth]{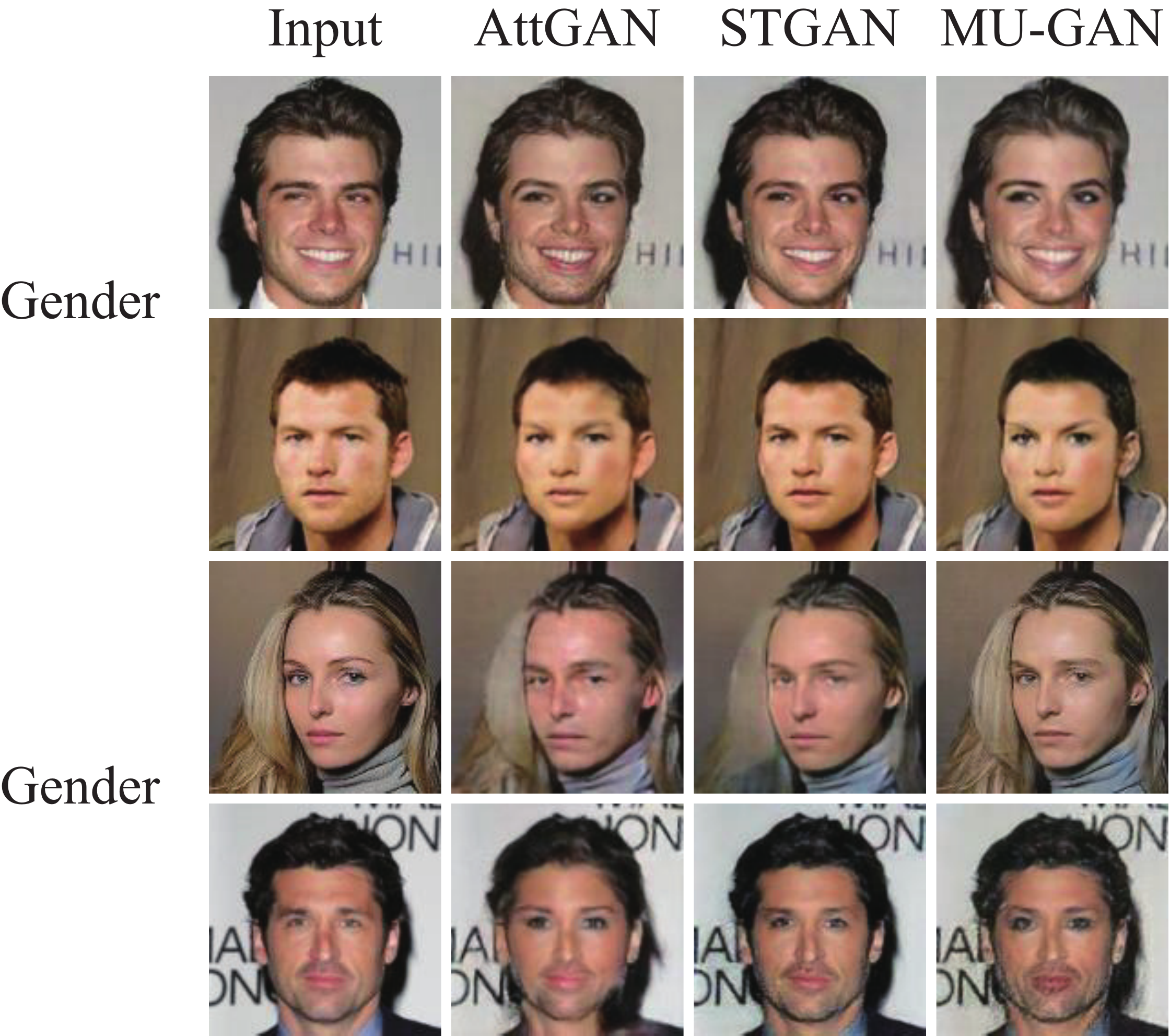}
			\label{7-c}
		\end{minipage}
	}%
	\centering
	\caption{ Attribute editing results at different abstract-level,compared with the competition methods.}
	\label{7}
\end{figure*}

First of all, from Fig. \ref{7-a}, our method outperforms other models. Samples generated by MU-GAN have better realism and fidelity of details. However, the results of its competing methods appear to be over-smooth and blurry with artifacts to some extent. One possible reason is that our model adopts a symmetrical U-Net-like architecture to make encoder representations complementary to decoder ones, without reducing its attribute editing ability. Besides, in a symmetric encoder-decoder, the corresponding encoder representation and decoder one are highly correlated.

Secondly, from Fig. \ref{7-b}, when editing global attributes, $e.g.$, \emph{Black Hair}, \emph{Blond hair}, \emph{Brown Hair}, and \emph{Pale Skin}, our method better enforces geometric constraints and is capable of distinguishing spatial regions related/unrelated to attributes, while its peers have difficulty in global attribute manipulation. For example, when the background is close to hair color, its peers often incorrectly recognize the background as hair, resulting in severe artifacts. In the opposite side, benefited from self-attention layers, our method can better distinguish the foreground and background, and accurately edits the hair color. In the same way, when dealing with a pale skin attribute, our method can better segment the faces from the background, rather than simply whitening the center region of an image, as done by its peers.

Thirdly, our model can effectively deal with the interference among attributes. Taking gender as an example, because of the sampling bias, the Male group generally have short-hair, while the Female group usually have long hair with neither beards nor mustache. Hair length, beard and mustache are attributes that are highly related to gender. As a result, these attributes often change with the editing of gender, which can be observed in generated results in the 3rd and 4th row of Fig. \ref{7-c}. The competition models sometimes drop beard attribute, when the image changes from male to female. These changes are very interesting and the generated samples are more realistic in attribute gender, but they can cause serious artifacts like fake long hair or make unexpected changes in other attributes. In our method, attributes are well decorrelated to avoid the interference among attributes and undesired changes in generated images.

\subsection{Quantitative evaluation}

\begin{figure}[hpb]
	\centering
	\includegraphics[width=0.5\textwidth]{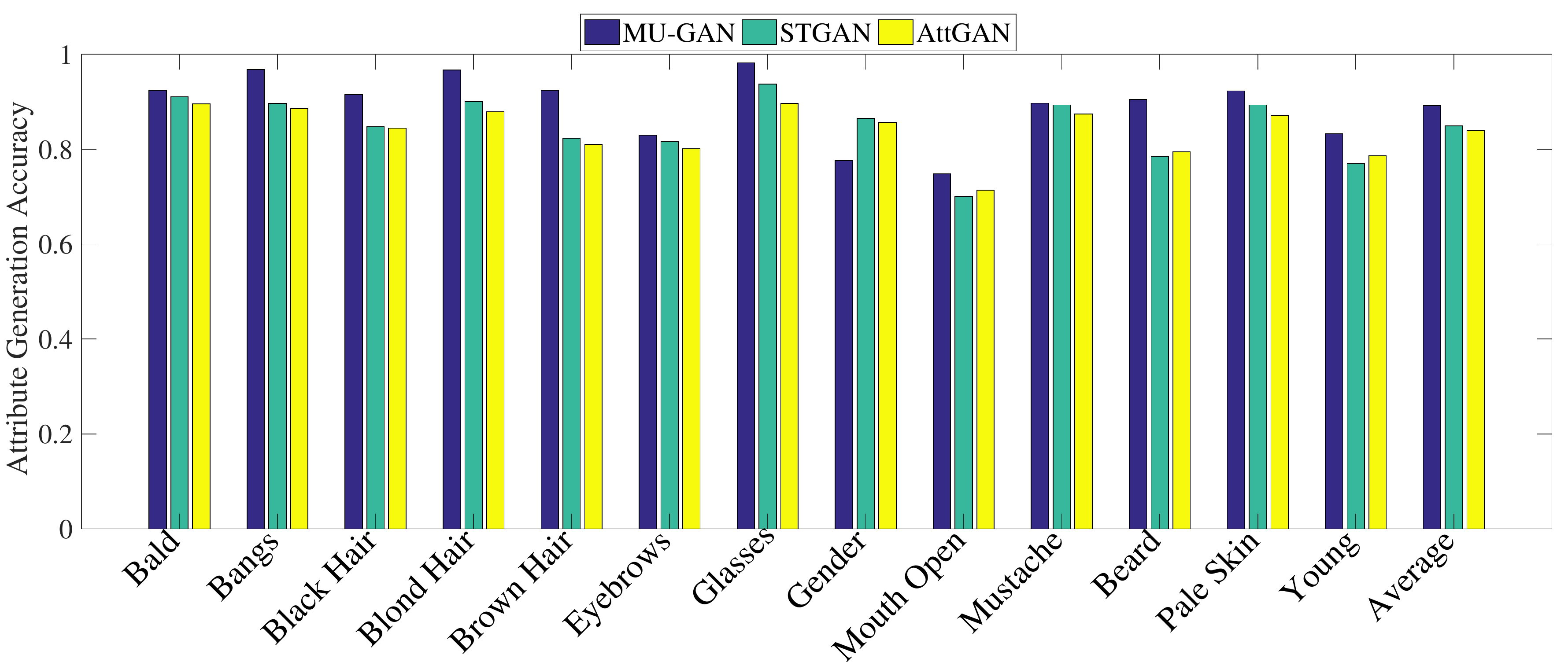}
	\caption{Attribute generation accuracy of AttGAN \cite{he2019attgan}, STGAN \cite{liu2019stgan} and MU-GAN.}
	\label{8}
\end{figure}

In a facial attribute editing task, the quality of generated images is mainly reflected in whether they are realistic or whether the source images are accurately manipulated from an original domain to a target one. We take attribute manipulation accuracy and reconstructed image quality for quantitative evaluation.
\begin{table}[h]\tiny
	\caption{Average attribute manipulation accuracy of the comparison methods on 13 facial attributes.}
	\label{to}       
	\centering
	\resizebox{0.5\textwidth}{!}{
		\begin{tabular}{lllllll}
			\hline\noalign{\smallskip}
			Method & ~~~ & AttGAN & ~~~ & STGAN & ~~~ & MU-GAN \\
			\noalign{\smallskip}\hline\noalign{\smallskip}
			Average Accuracy & ~~~ & 83.91\% & ~~~ & 84.89\% & ~~~ & 89.15\% \\
			\noalign{\smallskip}\hline
	\end{tabular}}
\end{table}
In order to evaluate the former, a multi-class classification method is employed to classify the generated images. First, a specific ResNet variant \cite{he2016deep} is trained on the training set of CelebA, attaining an accuracy of $94.79\%$ for 13 attributes on the test set. The classification network consists of three residual groups ($3$, $4$, $6$) and a fully connected layer with an output dimension of 13. The attribute generation accuracy is shown in Fig. \ref{8}. The classification results show that our method outperforms the others in the accuracy of attribute editing. As shown in Table \ref{to}, the average attribute generation accuracy of MU-GAN is $89.15\%$, which is a significant improvement over AttGAN's $83.91\%$ and STGAN's 84.89\%. Except for the gender attribute, the classification accuracy of other attributes are better than its peers, especially for the beard, hair colors and eyeglass attributes.
As mentioned earlier, gender correlates with other attributes, and MU-GAN is good at attribute decoupling, which is an effective way to prevent unexpected changes when editing target attributes. For example, when an image changes from male to female, MU-GAN faithfully retains the original beard and other correlated attributes, which is easily misjudged by the classification network in quantitative experiments. As we can see from Fig. \ref{7-c}, competing methods are more likely to make visually significant but unexpected changes like modifying hair length with serious artifact and other gender-related attributes while samples generated by MU-GAN change only the attributes that need to be changed. The above results also illustrate that MU-GAN not only has better attribute manipulation accuracy, but also has good attribute decoupling capabilities.

\begin{figure*}[htpb]
	\centering
	\includegraphics[width=\textwidth]{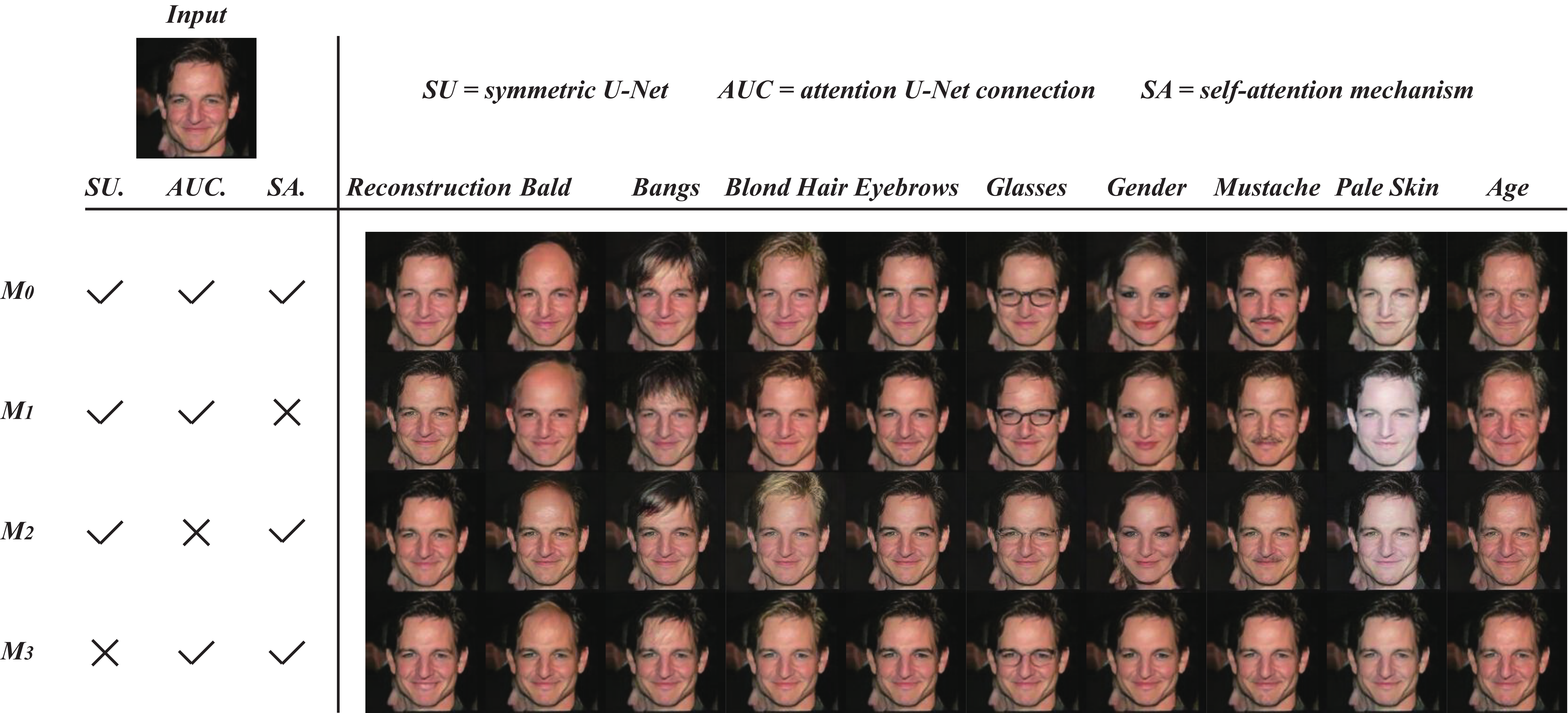}
	\caption{Effect of different combinations of the three components.}
	\label{13}
\end{figure*}

The evaluation indexes for reconstruction results are the Peak Signal to Noise Ratio and Structural SIMilarity (PSNR/SSIM). PSNR is the most common and widely-used evaluation index for images, but it is based on the error between corresponding pixel points, and it dose not take human visual characteristics into account. As a full-reference image quality evaluation index, SSIM measures image similarity in brightness, contrast, and structure. SSIM is better than PSNR in image denoising and similarity evaluation. In order to study the reconstruction ability, reconstruction image $\bm{x}^{\hat{a}}$ is generated from source image $\bm{x}^a$, conditioned on a source attribute vector. Table \ref{t1} lists the PSNR/SSIM results of reconstruction images for six methods. The quantitative results are consistent with the previous qualitative results \cite{liu2019stgan}. From Table \ref{t1}, benefited from AUCs, a symmetrical architecture, and a self-attention mechanism, our method can retain more image information and achieve much better reconstruction results than its five peers. AUCs are capable of generating high quality reconstruction results, which are more natural and realistic while retaining more details.
\begin{table}[t]\huge
	\caption{Reconstruction quality on facial attribute editing tasks.}
	\label{tab:1}       
	\centering
	\resizebox{0.5\textwidth}{!}{
		\begin{tabular}{lllllll}
			\hline\noalign{\smallskip}
			Method & IcGAN & FaderNet & StarGAN & AttGAN & STGAN & MU-GAN \\
			\noalign{\smallskip}\hline\noalign{\smallskip}
			PSNR/SSIM & 15.28/0.43 & 30.62/0.908 & 22.8/0.819 & 24.07/0.841 & 31.67/0.948 & 32.53/0.962\\
			\noalign{\smallskip}\hline
			\label{t1}
	\end{tabular}}
\end{table}

\section{Ablation Study}
In this section, we evaluate the effect of the two main components, $i.e.$, symmetric attention U-Net and self-attention mechanism on MU-GAN's performance. To analyze each's effect, we try different generator structures. Several MU-GAN variants are constructed, which are trained and tested in CelebA, under the same experimental settings. Ablation experiments between variants can also help to find out the contributions of AUCs, symmetric U-Net architecture, and self-attention mechanism.

We consider four variants: 1) $M_0$: original MU-GAN. 2) $M_1$: $M_0$ after removing a self-attention mechanism and only retaining a symmetrical attention U-Net architecture. 3) $M_2$: $M_0$ after removing AUCs, and retaining a symmetric encoder-decoder and a self-attention mechanism. 4) $M_3$: $M_0$ with an asymmetric encoder-decoder architecture.

\subsection{Effect of symmetrical attention U-Net structure}
First, a comparison of the editing result between symmetric and asymmetric encoder-decoder architectures is shown in Fig. \ref{13}. Compared with $M_0$, the generated results of $M_3$ to some extent, are blurrier and the image details are over-smooth. In addition, qualitative experimental results in Fig. \ref{13} illustrates that the MU-GAN variants with a symmetric encoder-decoder achieves better perceptual results on reconstructed images. One possible reason is that the symmetrical architecture avoids latent representation being highly-compressed, caused by the sharp decrease in the number of decoder channels, which can effectively retain the details and make edited results more natural and realistic.

\begin{figure}[h]
	\centering
	\includegraphics[width=0.5\textwidth]{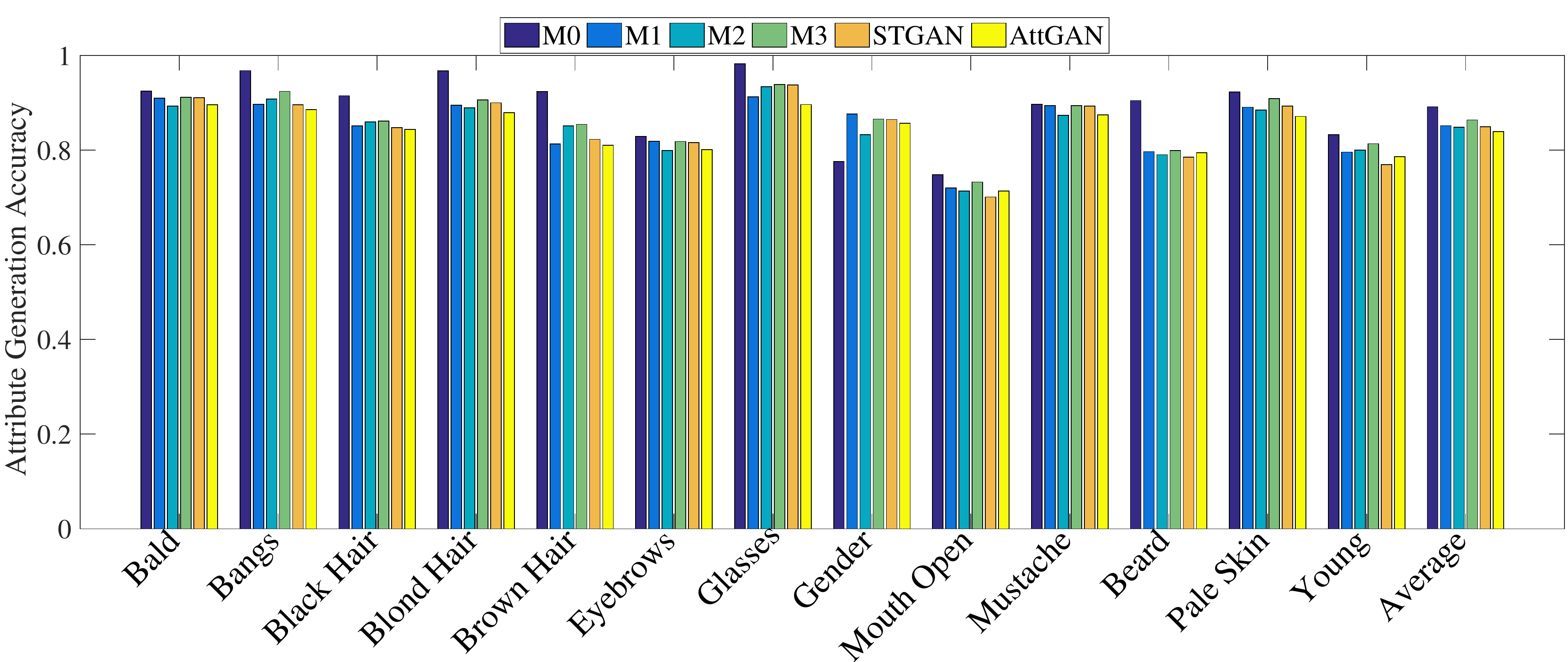}
	\caption{Attribute generation accuracy of MU-GAN variants, AttGAN, and STGAN, which are conducted on CelebA.}
	\label{9}
\end{figure}

Second, compared with models without AUCs ($e.g.$, $M_2$, AttGAN, and STGAN), $M_0$ has better attribute manipulation accuracy and higher PSNR/SSIM from Table \ref{t2} and Fig. \ref{9}. The additive attention mechanism selectively transfers attribute-irrelevant presentation from an encoder, filtering out the original attribute information to resolve the problems of information redundancy. Therefore, AUCs fuse multi-level features and enrich image details, which guarantees the attribute manipulation ability of AUC-based variants and only change the attributes that need to be changed.
\begin{table}[h]\huge
	\caption{Reconstruction quality and average classification accuracy of the MU-GAN variants.}
	\centering
	\resizebox{0.5\textwidth}{!}{
		\begin{tabular}{lllllll}
			\hline\noalign{\smallskip}
			Method & AttGAN & STGAN & $M_1$ & $M_2$ & $M_3$ & MU-GAN($M_0$) \\
			\noalign{\smallskip}\hline\noalign{\smallskip}
			PSNR/SSIM & 24.07/0.841 & 31.67/0.948 & 28.14/0.918 & 25.10/0.863 & 30.06/0.925 & 32.53/0.962\\
			\noalign{\smallskip}\hline\noalign{\smallskip}
			Accuracy & 83.91\% & 84.89\% & 85.15\% & 84.84\% & 86.36\% & 89.15\% \\
			\noalign{\smallskip}\hline
			\label{t2}
	\end{tabular}}
\end{table}

In addition, we have established many variants, which are based on symmetrical encoder-decoder model without self-attention mechanism to explore the effect of the number of AUCs on the results. $AUC_i$ means adding AUCs to the first $i$ layers, and $AUC_4$ is completely equivalent to $M_1$ mentioned earlier. As can be seen from Table \ref{t4}, the reconstruction quality and classification accuracy of the model are improved with the increase in the number of AUCs. When there are four AUCs in Generator, $AUC_4$ attains the best classification accuracy of $85.15\%$ and PSNR/SSIM increases from $24.07/0.841$ to $28.14/0.918$, which is a big breakthrough compared with the baseline. Therefore, we add AUCs to each layer of the generator. 

AttGAN's sparse encoder-decoder over-compresses image information, and loses a large number of details, which leads to low image fidelity. The introduction of skip-connections is one way to increase detail retention ability, but at the cost of severely weakening the attribute operation ability. Relevant/irrelevant information is indiscriminately injected into the decoder, resulting in information redundancy. With the help of the additive attention mechanism, AUCs can obtain the detailed information needed for image reconstruction. Similar to STGAN, we are committed to selectively transferring useful representation from encoder to decoder. AUCs avoid information redundancy, and then achieve the goal of balancing the detail retention ability and attribute operation ability, simultaneously.
\begin{table}[h]\huge
	\caption{The influence of AUCs on reconstruction quality and classification accuracy. There are MU-GAN variants with different amounts of AUCs.}
	\centering
	\resizebox{0.5\textwidth}{!}{
		\begin{tabular}{llllll}
			\hline\noalign{\smallskip}
			Method & AttGAN & $AUC_1$ & $AUC_2$ & $AUC_3$ & $AUC_4$($M_1$) \\
			\noalign{\smallskip}\hline\noalign{\smallskip}
			PSNR/SSIM & 24.07/0.841 & 26.10/0.862 & 27.43/0.880 & 27.89/0.891 & 28.14/0.918 \\
			\noalign{\smallskip}\hline\noalign{\smallskip}
			Accuracy & 83.91\% & 84.72\% & 85.13\% & 84.97\% & 85.15\%  \\
			\noalign{\smallskip}\hline
			\label{t4}
	\end{tabular}}
\end{table}

\subsection{Effect of a self-attention mechanism}
As shown in Fig. \ref{13}, attribute edited results generated by variants with self-attention mechanism, $e.g.$, $M_0$, $M_2$, and $M_3$, better enforce structural constraints, which can generate visual-reasonable results with rigorous geometry. Especially, our model performs well in global attribute editing such as \emph{bald} and \emph{pale skin}. Although the model with a self-attention mechanism can hardly obtain a significant improvement in the quantitative results, the generated images are more realistic in perception.

Benefited from a multi-attention mechanism, $M_0$ has strong attribute decoupling ability in abstract semantics. For example, in the task of gender attribute editing, the model with a multi-attention mechanism avoids unexpected changes. Attribute decoupling ability makes the gender attribute manipulation less significant compared with its peers, leading to quantitative classifier misjudgment.

To explore the effect of a self-attention mechanism, self-attention layers are added into different layers of the generator, which is based on symmetrical encoder-decoder model without AUCs. Since the effect of SA is mainly reflected in improving the quality of reconstructed images and has little effect on the improvement of classification accuracy, only the reconstruction experiment has been done here. As shown in Table \ref{t3}, self-attention mechanism seems to be more effective on the high-level feature graph $i.e.$, $Feat_{32}$ and $Feat_{64}$, but has limited performance improvement in the low-level graph. $Feat_{8,16}$'s performance is even lower than $Feat_{32}$. Theoretically, introducing self-attention mechanism into all layers of a generator is better. However, it will greatly increase the number of parameters of the model. Limited by hardware resources, we choose to add the self-attention layers on the third and fourth layers.

\begin{table}[h]\huge
	\caption{Comparison of MU-GAN variants, whose self-attention layer is placed in different positions. $Feat_i$ means adding self-attention layer to the $i \times i$ representation maps.}
	\centering
	\label{t3}
	\resizebox{0.5\textwidth}{!}{
		\begin{tabular}{|c|c|c|c|c|c|c|c|}
			\hline
			\multirow{2}*{Model} & \multirow{2}*{MU-GAN without SA} & \multicolumn{6}{c|}{MU-GAN with SA}\\
			\cline{3-8} 
			& & $Feat_8$ & $Feat_{16}$ & $Feat_{32}$ & $Feat_{64}$ & $Feat_{8,16}$ & $M_2(Feat_{32,64})$\\ \hline
			PSNR & 24.07 & 24.32 & 24.65 & 25.03 & 24.88 & 24.83 & 25.10\\ \hline
			SSIM & 0.841 & 0.845 & 0.848 & 0.859 & 0.852 & 0.850 & 0.863\\ \hline
		\end{tabular}
		}
\end{table}

\section{Conclusion}
The conclusion goes here.
In this paper, we introduce a multi-attention mechanism, $i.e.$, AUCs and self-attention mechanism into a symmetrical U-Net-like architecture, thus resulting in MU-GAN. By using AUCs, it can accurately edit desired facial attributes, which not only significantly improves attribute editing accuracy, but also enhances the detail retention ability. Furthermore, self-attention is introduced as a supplement to the convolutional layers and helps us generate results to better meet structural constraints. Experimental results prove that our method balances attribute manipulation and detail retention, and has strong decoupling capabilities. It can generate high-quality facial attribute editing results and outperforms the state-of-the-art approaches in terms of reconstruction quality and attribute generation accuracy. As future work, we intend to explore more appropriate attention mechanisms for AUCs to enhance the performance of MU-GAN.


%

\appendices
%

\section*{Acknowledgment}
The authors gratefully acknowledge the support of NVIDIA Corporation with the donation of the GPU used for this research.

\ifCLASSOPTIONcaptionsoff
  \newpage
\fi

\small
\bibliographystyle{IEEEtran}
\bibliography{MU-GAN}

\begin{IEEEbiography}[{\includegraphics[width=1in,height=1.25in,clip,keepaspectratio]{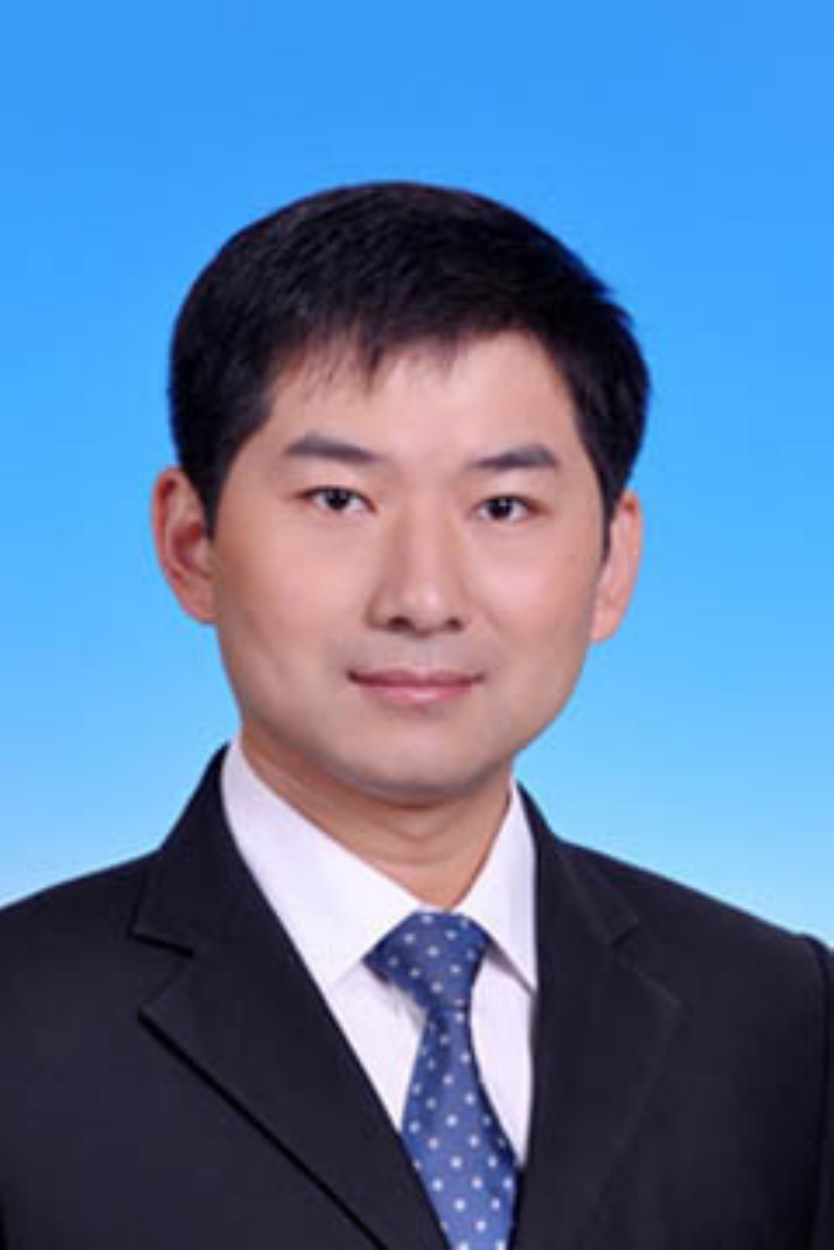}}]{Ke Zhang}
	received his M.E. degree in signal and information processing from North China Electric Power University, Baoding, China, in 2006, and the Ph.D. degree in signal and information processing from the Beijing University of Posts and Telecommunications, Beijing, China, in 2012. He finished his Post Doc in computer vision from the University of Missouri, Columbia, MO, USA, in 2016. He is currently an Associate Professor with North China Electric Power University. His research interests include computer vision, deep learning, machine learning, robot navigation, natural language processing, and spatial relation description.
\end{IEEEbiography}
\vspace{-5pt}
\begin{IEEEbiography}[{\includegraphics[width=1in,height=1.25in,clip,keepaspectratio]{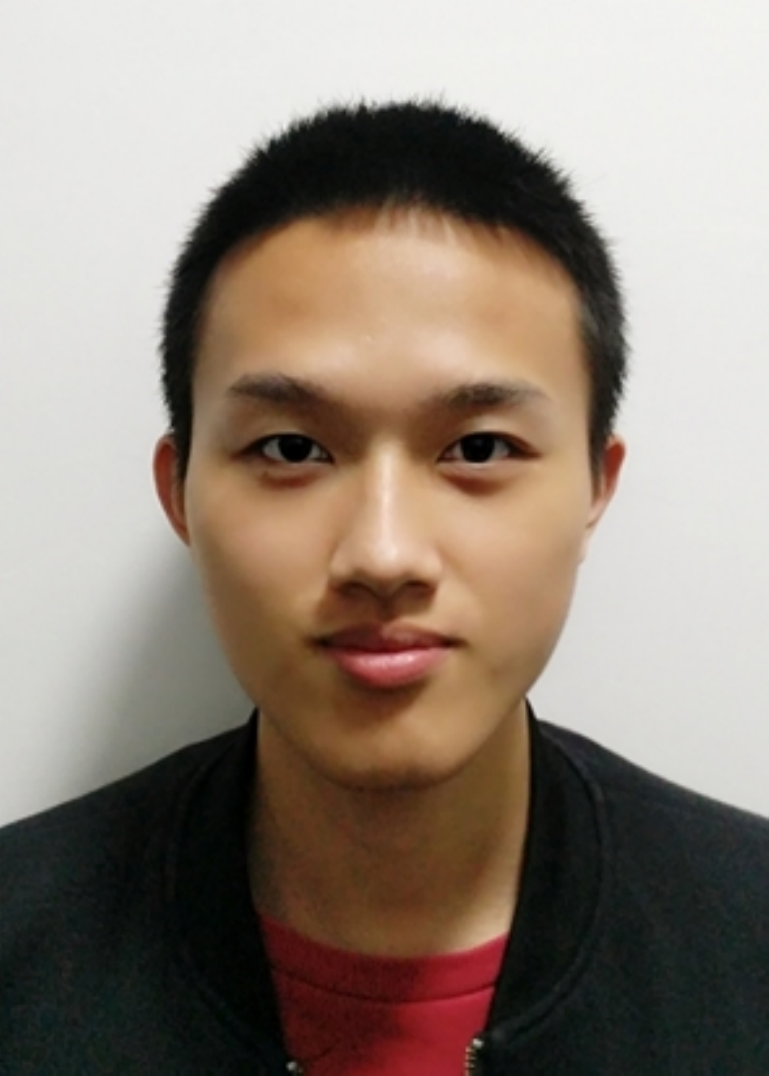}}]{Yukun Su}
	received his B.S. degree in electronic information science and technology from North China Electric Power University, Baoding, China, in 2018, where he is currently pursuing the master's degree in communication and information engineering. His research interests include computer vision and Generative Adversarial Networks.
\end{IEEEbiography}
\vspace{-5pt}
\begin{IEEEbiography}[{\includegraphics[width=1in,height=1.25in,clip,keepaspectratio]{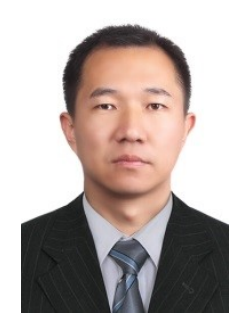}}]{Xiwang Guo}
	received his B.S. degree in Computer Science and Technology from Shenyang Institute of Engineering, Shenyang, China, in 2006, M.S. degree in Aeronautics and Astronautics Manufacturing Engineering. from Shenyang Aerospace University, Shenyang, China, in 2009, Ph. D. degree in System Engineering from Northeastern University, Shenyang, China, in 2015. He is currently an associate professor of the College of Computer and Communication Engineering at Liaoning Shihua University. From 2016 to 2018, he was a visiting scholar of Department of Electrical and Computer Engineering, New Jersey Institute of Technology, Newark, NJ, USA. He has published 40+ technical papers in journals and conference proceedings, including IEEE Transactions on Cybernetics, IEEE Transactions on System, Man and Cybernetics: Systems, IEEE Transactions on Intelligent Transportation Systems, and IEEE/CAA Journal of Automatica Sinica. His current research interests include Petri nets, remanufacturing, recycling and reuse of automotive, intelligent optimization algorithm.
\end{IEEEbiography}
\vspace{-5pt}
\begin{IEEEbiography}[{\includegraphics[width=1in,height=1.25in,clip,keepaspectratio]{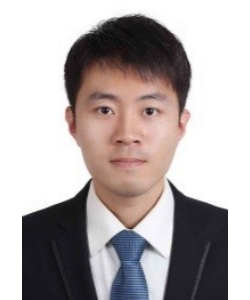}}]{Liang Qi}
	received his B.S. degree in Information and Computing Science and M.S. degree in Computer Software and Theory from Shandong University of Science and Technology, Qingdao, China, in 2009 and 2012, respectively, and Ph.D. degree in Computer Software and Theory from Tongji University, Shanghai, China in 2017. He is currently with Computer Science and Technology at Shandong University of Science and Technology, Qingdao, China. From 2015 to 2017, he was a visiting student in the Department of Electrical and Computer Engineering, New Jersey Institute of Technology, Newark, NJ, USA. He has published over 40 technical papers in journals and conference proceedings, including IEEE Transactions on System, Man and Cybernetics: Systems, IEEE Transactions on Intelligent Transportation Systems, and IEEE/CAA Journal of Automatica Sinica. He received the Best Student Paper Award-Finalist in the 15th IEEE International Conference on Networking, Sensing and Control (ICNSC’2018). His current research interests include Petri nets, machine learning, optimization, and intelligent transportation systems.
\end{IEEEbiography}
\vspace{-5pt}
\begin{IEEEbiography}[{\includegraphics[width=1in,height=1.25in,clip,keepaspectratio]{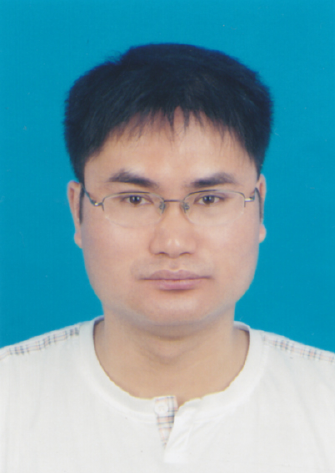}}]{Zhenbing Zhao}
	received his B.S., M.S., and Ph.D. degrees from North China Electric Power University, Baoding, in 2002, 2005, and 2009, respectively. He is currently an Associate Professor with North China Electric Power University. His research interests include machine learning, image processing, and the intelligent detection of electrical equipment.
\end{IEEEbiography}

\end{document}